# Global-to-local image quality assessment in optical microscopy via fast and robust deep learning predictions

Elena Corbetta[a,b], Thomas Bocklitz[a,b]*

a. Leibniz Institute of Photonic Technology, Member of Leibniz Health Technologies, Member of the Leibniz Centre for Photonics in Infection Research (LPI), Albert-Einstein-Strasse 9, 07745 Jena, Germany.
b. Institute of Physical Chemistry (IPC) and Abbe Center of Photonics (ACP), Friedrich Schiller University Jena, Member of the Leibniz Centre for Photonics in Infection Research (LPI), Helmholtzweg 4, 07743 Jena, Germany

*Corresponding author

E-mail address: thomas.bocklitz@uni-jena.de

## Abstract

Optical microscopy is one of the most widely used techniques in research studies for life sciences and biomedicine. These applications require reliable experimental pipelines to extract valuable knowledge from the measured samples and must be supported by image quality assessment (IQA) to ensure correct processing and analysis of the image data. IQA methods are implemented with variable complexity. However, while most quality metrics have a straightforward implementation, they might be time consuming and computationally expensive when evaluating a large dataset. In addition, quality metrics are often designed for well-defined image features and may be unstable for images out of the ideal domain. To overcome these limitations, recent works have proposed deep learning-based IQA methods, which can provide superior performance, increased generalizability and fast prediction. Our method, named μDeepIQA, is inspired by previous studies and applies a deep convolutional neural network designed for IQA on natural images to optical microscopy measurements. We retrained the same architecture to predict individual quality metrics and global quality scores for optical microscopy data. The resulting models provide fast and stable predictions of image quality by generalizing quality estimation even outside the ideal range of standard methods. In addition, μDeepIQA provides patch-wise prediction of image quality and can be used to visualize spatially varying quality in a single image. Our study demonstrates that optical microscopy-based studies can benefit from the generalizability of deep learning models due to their stable performance in the presence of outliers, the ability to assess small image patches, and rapid predictions.

# 1. Introduction

As advanced microscopy techniques reveal deeper insights in the principles governing cellular and subcellular processes, the importance of good quality images increases. Image quality has a significant impact on image analysis and tissue diagnosis, determining correct or wrong outcomes. The enhancement of image quality does not involve only a visual perception issue for better aesthetics but includes also all the strategies for recognizing and correcting experimental artifacts and deviations of a measurement from the actual sample content. Image quality assessment (IQA) is the standard term referring to tools and processes for evaluating the goodness of the images. IQA also determines whether further steps are needed for quality improvement, including adjustment of measurement parameters and image restoration tasks. Therefore, IQA is an essential validation step for every new method proposed by researchers, as the good performance of experimental techniques and computational methods must be demonstrated through a quantitative assessment. This often involves benchmarking new approaches against established techniques.

IQA can be executed with a variety of strategies, which are selected depending on prior information available and the definition of quality for the current evaluation task. Depending on the knowledge of the ideal image, we refer to full-reference IQA when the ground truth is available, reduced-reference IQA when limited information on the target quality is available, and no-reference IQA when no prior information is available.

In addition, different methods can be grouped according to their increasing complexity and the reduced requirement of user contribution, as schematized in Figure 1. The simplest workflow involves the definition of one or multiple quality metrics that are computed and manually inspected to determine trends within the dataset and select the images with the preferred features (Figure 1 (a)). This method leaves great control to the user, that can tune the evaluation and carefully analyse the results. However, it implies longer evaluation time and in-depth knowledge of the quality metrics.

The metrics used for manual evaluation can be used as starting point for automating IQA (Figure 1 (b)). The metrics can be used to generate a comprehensive model for image quality, which should properly combine the metrics to classify the images based on their quality or generate an overall quality score. Recent studies have described machine learning (ML)-based methods which offer also a good degree of interpretability thanks to the straightforward implementation and the use of known metrics.[1] These methods greatly automate IQA and increase objectivity in the evaluation but still require the computation of parameters for every image.

The evaluation task can be further sped up by implementing deep learning (DL)-based approaches to estimate quality features or quality scores directly from the images. These methods require a reliable quality score to be defined for the training dataset, as well as the use of complex architectures to extract hidden features from the images. However, they have great potential for generalization and regularization of IQA in the presence of outliers, as wells as speeding up the evaluation task. These advantages have been demonstrated for natural images, but the implementation of DL-based IQA is still rare in optical microscopy and biomedical fields. Recent studies have explored the application of DL for assessing image focus in optical microscopy[2, 3], blindly evaluating image distortions in pathological images[4], and for no-reference IQA of confocal endoscopy images based on perceptual local descriptors[5].

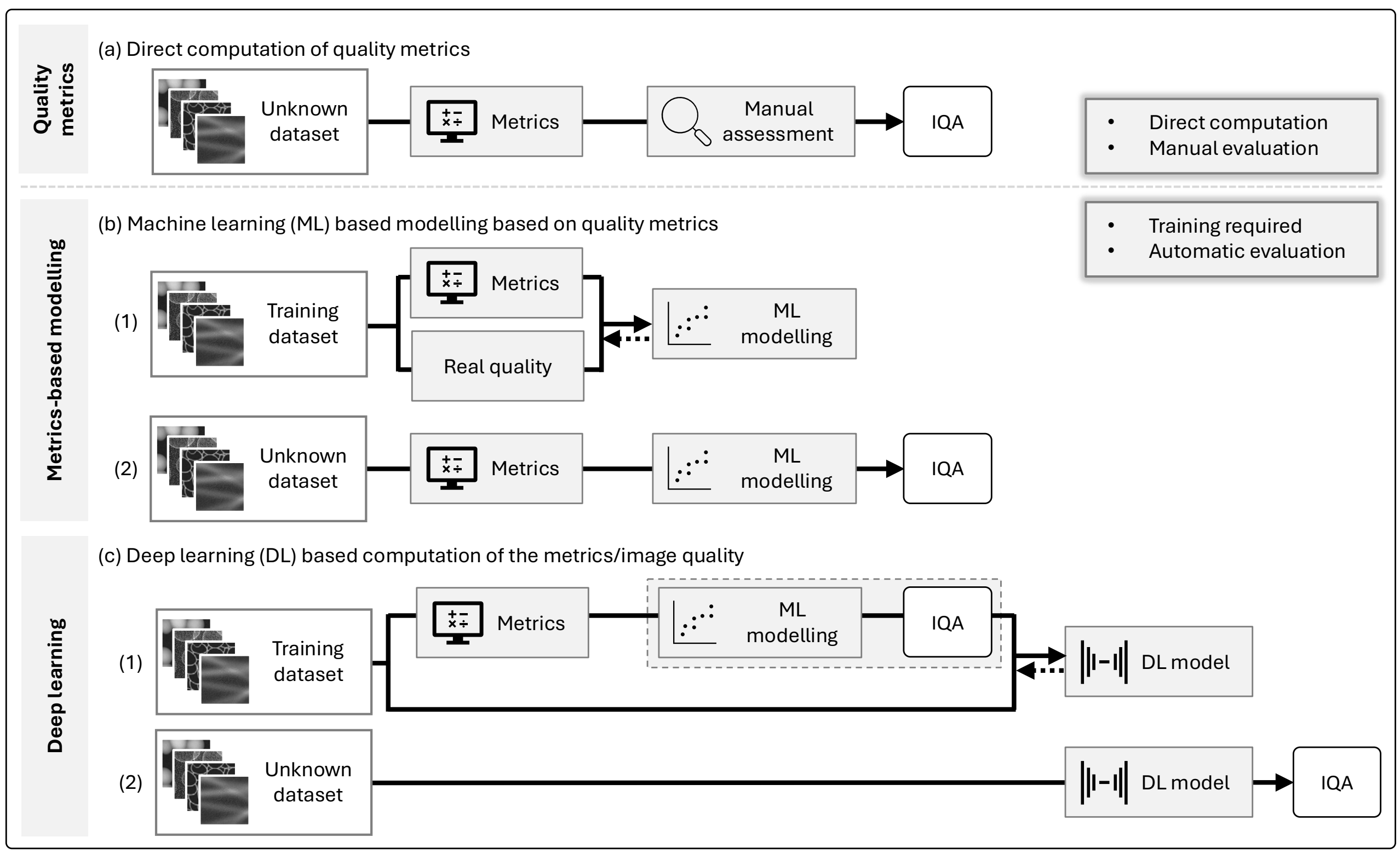

*Figure 1 - **Image quality assessment (IQA) for optical microscopy: schematics of standard methods and deep learning (DL)-based approaches.** (a) Standard methods rely on the computation of quality metrics or similarity metrics. The final assessment is obtained prior manual inspection of the results to determine the best image. (b) The first step towards an automatic IQA is the integration of the metrics into machine learning (ML)-based methods: the metrics are computed for a known dataset and used to model the image quality during the training process. After training, automatic computation of a quality score can be obtained, thus eliminating the manual assessment step required by the first approach. (c) DL-based methods can be used to skip the computation of the metrics and obtain IQA scores directly from the images. During the training process, the DL model learns to predict single quality metrics or the ML-generated quality score (dashed frame). Then, the trained DL model can be used for automatic IQA for unknown datasets.*

In optical microscopy, robust protocols for IQA are based on standard metrics[6, 7], while DL-based methods are largely applied to quality improvement, such as image super-resolution, image reconstruction, and denoising. [8, 9] Regardless the low amount of works dedicated specifically to DL-based IQA, the good performance of methods for natural image encourages for further investigation.[10-12]

In this paper, we present a Convolutional Neural Network (CNN) for image quality assessment in microscopy, named µDeepIQA, inspired to the WaDIQaM no-reference method proposed by Bosse et al.[10, 13] for natural images, and based on the VGG architecture[14]. We adapted the previous method to fluorescence microscopy applications and demonstrated that the model can be trained and applied to the estimation of single quality features and comprehensive quality scores, which have been modelled originally from multiple metrics. This study demonstrates the good performance of the method compared to the state of the art and the substantial acceleration of IQA. Moreover, we show that µDeepIQA regularizes IQA for image outliers with uncommon features, for which the computation of standard metrics fails; this advantage ensures fully automated IQA for large datasets and avoids excessive tuning when evaluating outputs from different experimental conditions or processing algorithms. Moreover, µDeepIQA predicts quality on small patches and therefore allows local, targeted IQA, an interesting feature that was already explored by previous studies[3]. This feature is not often available for standard metrics, which require a bigger number of pixels for a proper evaluation. Our method boosts IQA in optical microscopy, allowing for a fast, local, robust evaluation of images with a variety of experimental artifacts, without the need of fine-tuning depending on the specific measurement. This is particularly important for biomedical studies, where the measurements might be affected by variable artifacts within the field of view (FOV), and for successful analysis and diagnosis.

# 2. Results and discussion

## 2.1 Prediction of image resolution and quality scores with µDeepIQA models

The DL method presented in this paper is based on the architecture introduced by Bosse et al.[10, 13], which was successfully tested on natural images. µDeepIQA is inspired by the no-reference architecture with weighted patch aggregation of predicted image quality. The detailed implementation is shown in Figure 2 (a,b) and described in the Methods section. Briefly, the CNN model receives as input small patches of single-channel images and extracts features with a contractive CNN structure; then, two parallel feature regression branches estimate the single-patch quality $y_i$ and single-patch weight $\alpha_i$, or salience; finally, the image quality $y$ is predicted as weighted average of single-patch predictions (Equation (1)). This approach is particularly relevant for microscopy applications because it ensures that each patch contribution is weighted according to its content. Indeed, a background patch that contains no imaged structure might not be as relevant to the quality prediction of a high-quality image. On the other hand, if there is strong background noise, the same patch might be more relevant for the final estimate.

The same CNN architecture can be trained to estimate any quality metrics. In this manuscript, we show the prediction of two parameters that are often investigated in microscopy-based studies: the optical resolution and a comprehensive quality score. Therefore, two models are trained in a supervised manner on a semisynthetic dataset containing multiple experimental and simulated structures with 5 simulated artifacts of variable intensity and high-quality images, as explained in Methods. Supplementary Figure 1 provides an overview of the underlying clean samples. The true labels for training are obtained by estimating the image resolution with Fourier Ring Correlation ($FRC_{res}$), and predicting the image quality score through metrics-based modelling of image quality (as schematized in Figure 1 (b)) with the Multi-Marker (MM)-IQA method, published in our previous work[1].

Figure 2 (c, d) shows the loss function for the training and validation datasets until the selected number of training epochs (300), demonstrating good convergence of the model. Figure 2 (e, f) shows the target vs predicted $FRC_{res}$ and MM quality score for the test dataset: the artifacts affecting the images are visualized in different colours, *reference* indicates high quality images, and the marker shape indicates the underlying imaged structure, showing no bias to specific samples. The plots show good linear trend of the predictions. Remarkably, the deviations from linearity in the MM quality score plot (Figure 2 (f)) are due to a shift of reference and vignetted images to higher scores compared to the target value, and blurred images to a lower score. This beneficial regularization applied by the CNN is further inspected in the following paragraphs.

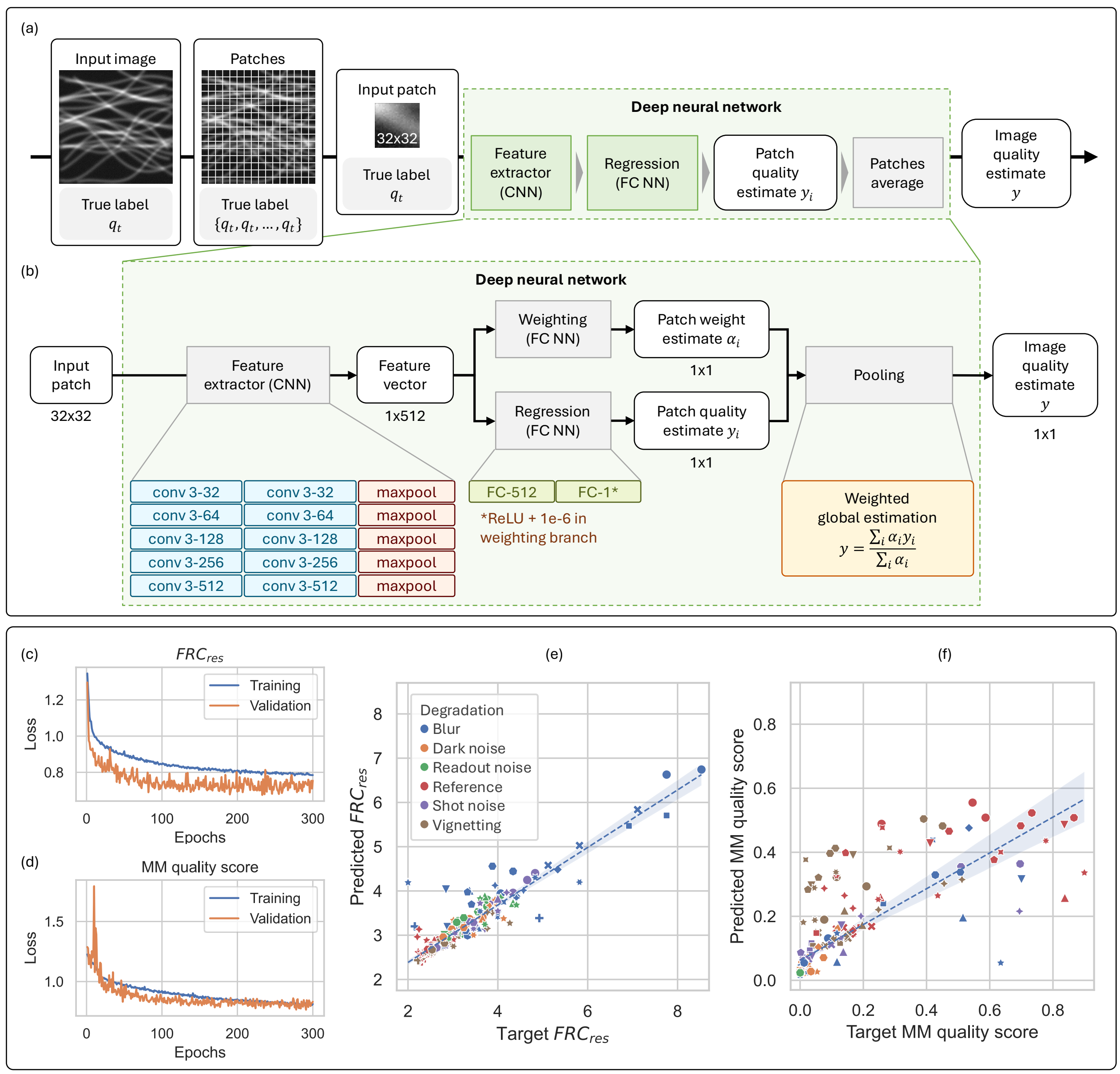

*Figure 2 -* ***µDeepIQA: convolutional neural network (CNN) and prediction of quality metrics and quality scores for a semisynthetic dataset.*** *(a) Workflow for DL-based IQA. Input images are split into patches of 32 x 32 pixels that are fed to the DL model. The network is composed by a CNN for feature extractor and a fully connected (FC) section that outputs the patch prediction. Then, a final averaging layer computes a weighted average of all patches to retrieve the global image quality estimate. The CNN is trained on semisynthetic images of multiple images of 15 different samples affected by 6 varying artifacts. (b) Detailed schematic of the CNN architecture. The feature extractor is composed by a series of 5 pairs of convolutional layers of varying size followed by a max pooling layer. Then, two parallel branches of fully connected (FC) layers estimate the patch quality $y_i$ and the patch weight $\alpha_i$. Finally, a weighted average generates the image quality estimate y. The same architecture is trained to generate models that can predict single quality metrics or global quality scores. (c, d) Training and validation loss function for a model that predicts image resolution estimated by Fourier Ring Correlation ($FRC_{res}$) (c), or a global quality score generated by metrics-based modelling (d). (e,f) Target vs predicted values for a model trained on $FRC_{res}$ (e) or a global quality score (f). Every point is an image of the test dataset: colours indicate the image artifact, and marker shapes the imaged sample, showing no bias towards the underlying image structure. The dashed blue line is the linear regression with confidence interval of 95% indicated by the shadowed region. The black dashed line is the identity line.*

## 2.2 µDeepIQA regularizes quality rankings in presence of non-standard artifacts

After assessing the successful training on a broad semisynthetic dataset with varying image quality, we assert the robustness of the model by pushing the predictions to non-standard artifacts that were not included in the training dataset.

Many quality metrics are developed to work with specific image features, for example a minimal amount of image noise. FRC estimates the image resolution as spectral signal-to-noise ratio and requires statistical intensity oscillations for a correct computation. Similarly, signal-to-noise ratio is not sensitive to a detrimental smoothing of the image if the signal intensity range remains the same. For these reasons, metrics-based IQA modelling might fail to predict a good MM score for smoothed and noise-free images, because the score is estimated directly from the metrics. On the contrary, the generalization capability of CNNs potentially resolves these issues in presence of atypical artifacts that were not included in the training dataset.

Figure 3 (a) shows the DL prediction and standard $FRC_{res}$ obtained for a set of 15 samples affected by 5 increasing levels of noise-free blurring, 5 levels of noise-free vignetting, and 5 low noise oscillations to generate independent high-quality images. The CNN predicts the best spatial resolution for high-quality images, slightly worse resolution for vignetted images, which are degraded only at the edges, and worsening resolution for blurred images. Instead, standard computation via FRC results in the best resolution assigned to highly blurred images due to high correlation in the Fourier domain in absence of noise. An initial worsening of the resolution is detected on average for the first 3 blur levels, but the increased smoothing is interpreted as a beneficial feature.

Figure 3 (b) shows the DL-predicted and metrics-based MM quality score for the same dataset. The CNN predicts the highest quality for high-quality images and low artifacts, with clearly decreasing score for strong blurring. Metrics-based modelling, instead, misinterprets the increasing smoothing and shows the opposite behaviour because the score is influenced, but not solely determined, by the $FRC_{res}$. Vignetting and high quality are well predicted in both cases. The MM score of vignetted images shows an ideal trend because vignetting prediction is weakly dependent on FRC and relies more on quality metrics that can be computed correctly in the absence of noise.

Figure 3 (c-e) show the quality rankings generated by the DL-predicted and metrics-based MM quality scores for the subset of blurred images. The images are labelled by a unique index and are coloured according to the true artifact level. The top bar represents the ideal distribution of the images in the ranking according to the known quality. As highlighted in Figure 3 (b), the blurred images score the best quality for the metrics-based modelling, but they are successfully shifted to lower positions by the CNN. Indeed, the DL-based ranking shows a distribution much more aligned to the ideal one and the generation of intermediate score that transition from the maximum to the minimum value less steeply.

The Kendall's rank correlation coefficient (KRCC) allows to estimate the degree of accordance between the ideal and predicted rankings (see Supplementary Figure 2). For the subset of blurred images, it results in a value of 1 (perfect correlation with the artifacts) for µDeepIQA and -0.2 for MM-IQA. For vignetted images, both methods achieve perfect correlation.

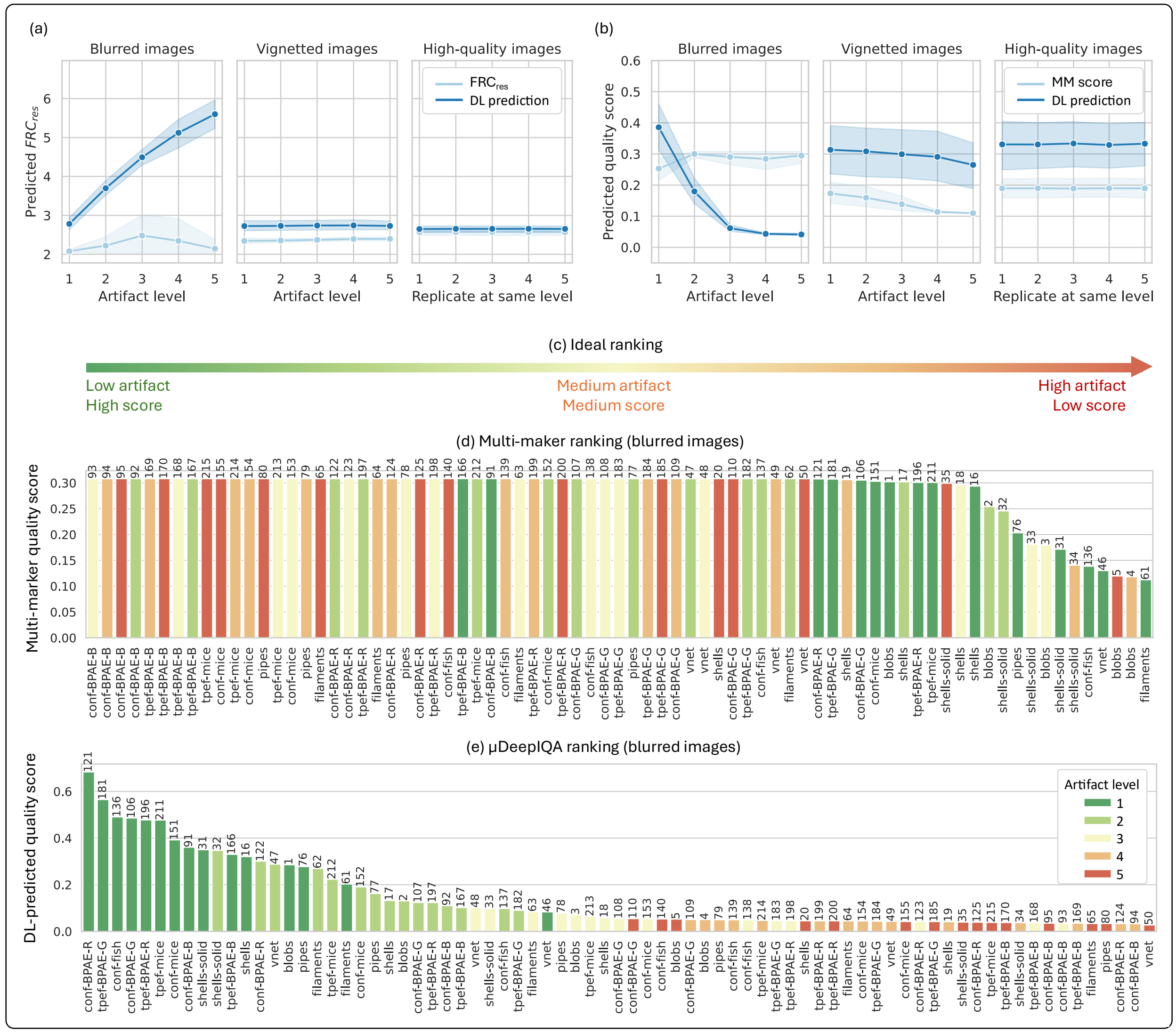

*Figure 3 - **Regularization of quality rankings by DL-based IQA.** (a) Resolution estimated by Fourier Ring Correlation obtained by direct computation ($FRC_{res}$) and predicted by the trained model (DL prediction) for sets of images with increasing blur, increasing vignetting, and high-quality images. Degraded images are noise-free: this impedes the correct FRC computation, but the DL model can predict the correct trend. (b) Quality score of the same images predicted by MM-IQA (MM score) and the trained model (DL prediction). For blurred images, MM-IQA prediction shows bad alignment with the real artifact due to the significant contribution of FRC in the prediction of blurring. Instead, µDeepIQA shows good performance for every case. (c) The quality scores predicted by MM-IQA and µDeepIQA are used to build quality rankings for the subset of blurred images. Rankings are color-coded according to the real artifact level: dark green for the first blurring level, and transition to yellow, orange and red for increasing blurring. The ideal ranking indicates the expected distribution of the images according to their true quality. (d) Ranking generated with the MM-IQA score, based on direct computation of the metrics: images with strong blurring are placed at top positions due to the detrimental contribution of the FRC value. (e) Ranking generated with µDeepIQA: DL predictions not only generate a more gradual transition of the quality score, but also regularize the position of low blurring levels, placing them at the top and middle positions. The height of the bar is the predicted quality score, the bar label is the unique image index, and the x axis shows the sample label (as defined in Supplementary Figure 1).*

## 2.3 Local patch-wise predictions of image quality

So far, we have investigated the prediction of image quality scores and the generation of quality rankings. As pointed out also in the original study on natural images, the weighted patch aggregation provides superior performance because it can dynamically recognize the most relevant patches and the quality of different image regions. While this tool has been proposed originally as internal optimization strategy, we exploit single patch predictions for local image quality assessment and demonstrate that local predictions are aligned with the spatial distribution of image artifacts.

µDeepIQA enables local IQA with patch resolution (32 x 32 pixels, in our model), which is not always achievable with state-of-the-art methods for microscopy. For example, FRC resolution and in general every investigation of the frequency components requires to build a statistic from larger images. In a similar way, the signal to noise ratio requires to capture the full signal range and possibly enough noisy oscillations, while µDeepIQA can evaluate in parallel single background patches and brighter, structured regions.

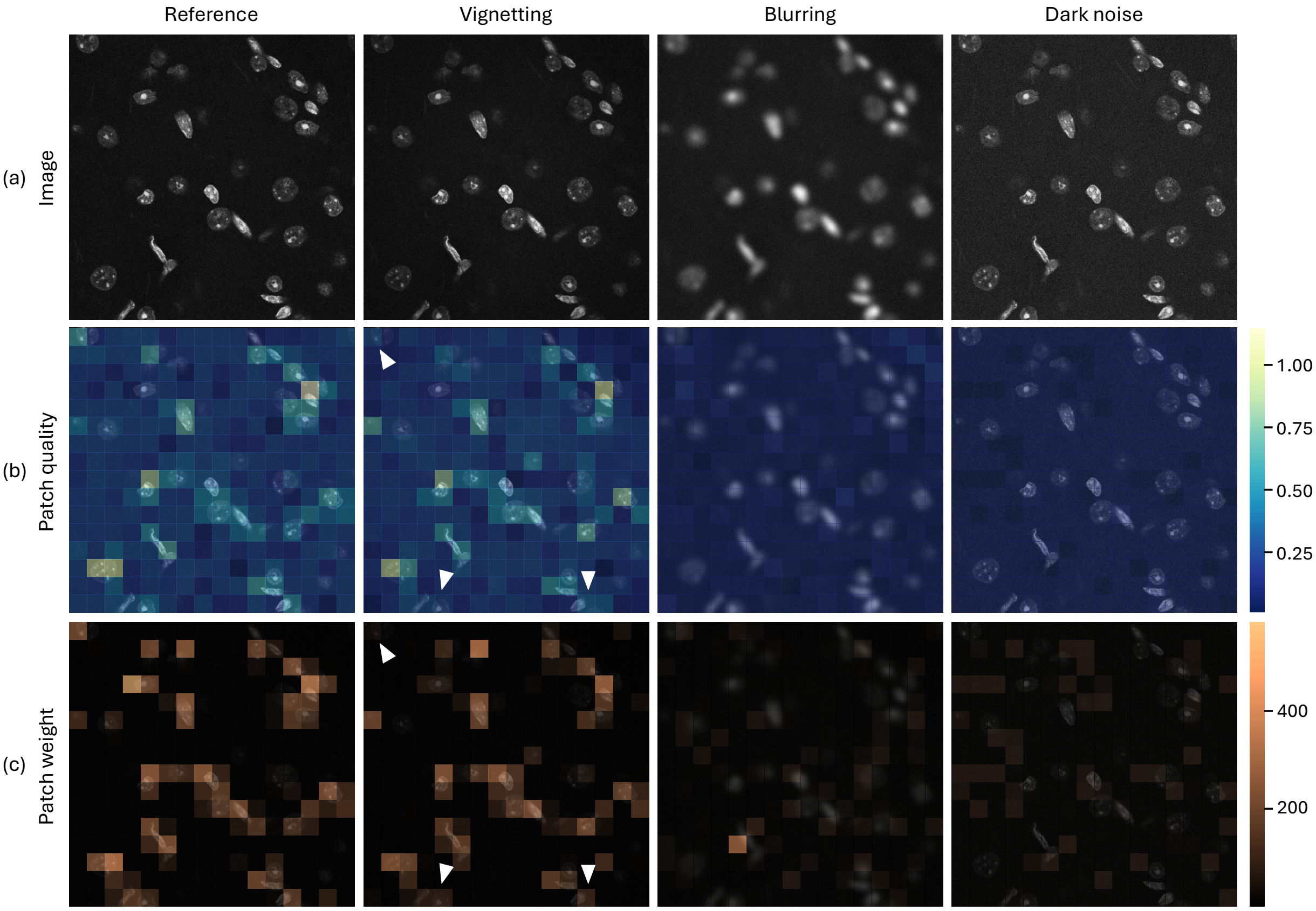

*Figure 4 –* ***Inspection of the patch quality predicted by µDeepIQA for different simulated artifacts.*** *The predicted image is a high quality two-photon microscopy measurement of fixed mouse brain tissues with four simulated artifacts. The headers of each column indicate the artifact, where reference is a high-quality with low noise level. (a) Predicted input images. (b) Single-patch quality overlayed on the input image. The predicted quality of reference and vignetted images depends on the single patch, with higher value assigned to patches containing bright objects, and the lower values for background patches. The vignetted image shows a lower quality at the edges, where the intensity decreases (see white arrowheads). Blurred and noisy images have low and uniform patch quality. (c) Single-patch weight overlayed on the input image. The predicted weight of reference and vignetted images depends on the single patch, similarly to the patch quality estimate, with a decrease of the patch weight at the edges of the vignetted image (see white arrowheads). Blurred and noisy images have uniform patch weights, indicating that the model identifies an artifact spread in the same way in the whole image. Patch size: 9,6 µm.*

Figure 4 shows the patch-wise quality and weight for four test images with varying artifacts. Figure 4 (a) shows the images and Figure 4 (b) the local patch-quality predictions. For high quality images (Reference, in the first column), µDeepIQA predicts higher patch quality in the location of imaged objects and low quality in the background. Instead, blurred and noisy images have a uniform lower quality overall. In the case of vignetting, that affects the quality only at the image edges, the decrease in patch quality is predicted only at image borders, as indicated by the white arrowheads. Similar results are obtained for patch saliency (Figure 4 (c)).

These findings give further insights on the correlation results. By inspecting KRCC for single samples (Supplementary Figure 2 (b)), we observe that μDeepIQA provides the most accurate predictions for blurred images, while the multi-marker approach is preferable for overall vignetting prediction. Indeed, μDeepIQA detects local artifacts and treats vignetted patches as background regions with low quality and low weight. Consequently, vignetted images receive a high score, determined solely by averaging the bright patches. On the other hand, MM-IQA detects vignetting and predicts global scores for the full images (Figure 3 (b)).

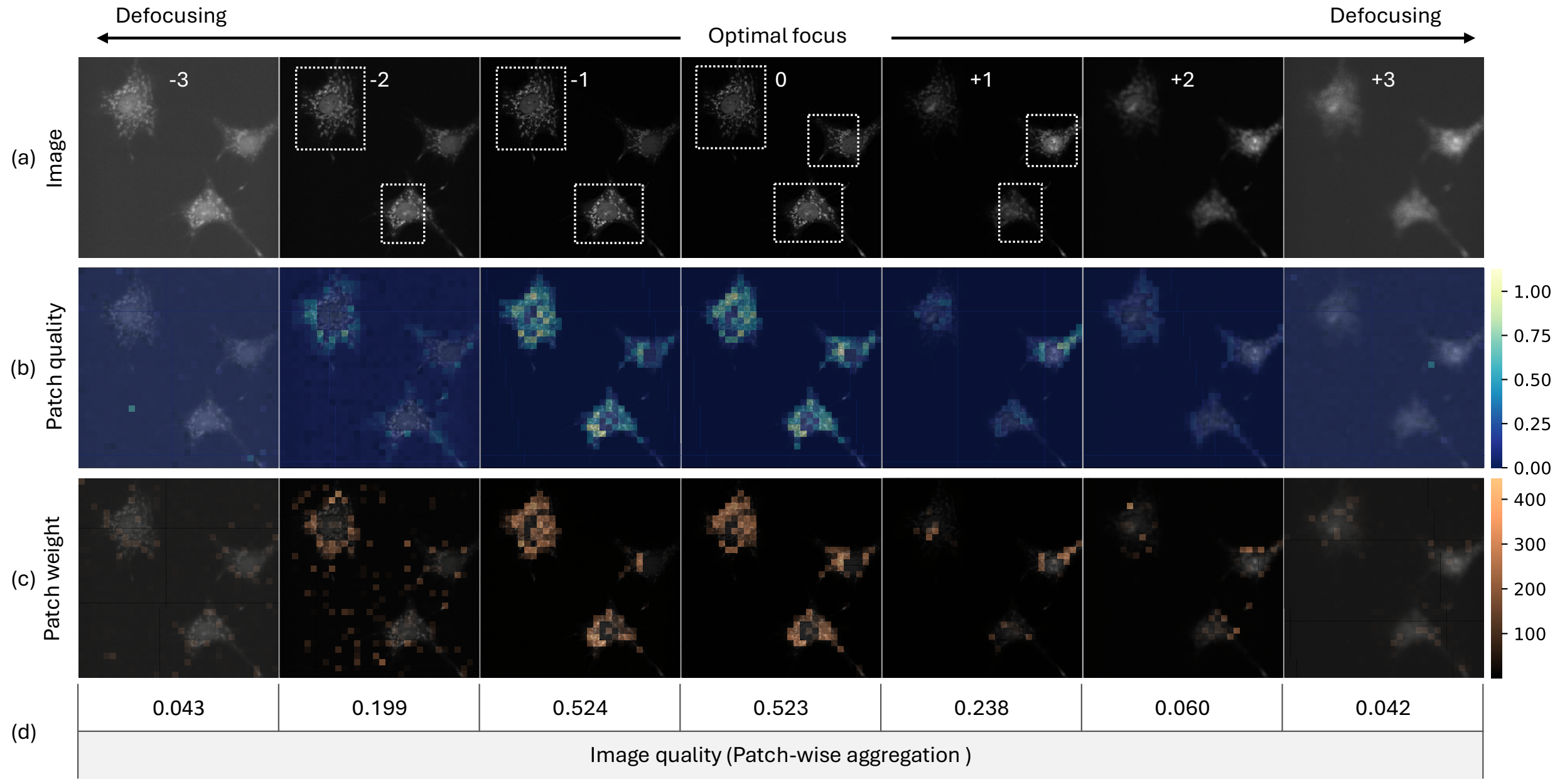

*Figure 5 – **Inspection of the patch quality and weight predicted by μDeepIQA for experimental measurements at different focal positions.** The predicted image is a high quality two-photon microscopy measurement of fixed bovine pulmonary artery endothelial cells (mitochondria). (a) Predicted input images. White labels indicate the position relative to the optimal focus, indicated as 0. Even though the optimal focus is the best position along the optical axis to image most of the structures, a few details are in focus in different images, as highlighted by white dashed frames. (b) Single-patch quality overlayed on the input images. The predicted quality is low and uniform far from the optical focus. For images ±1 and 0 the quality is lower in the background and higher on the structures in focus. We can observe how the value of the patch quality follows the focus change by highlighting in each position the most focused regions (for example, in the upper-right small cell). (c) Single-patch weight overlayed on the input images. The predicted quality is low and uniform for ±3 images while, getting closer to the optimal focus, it becomes lower in the background and higher on the imaged structures, similarly to the quality prediction. (d) Quality score predicted for each image after weighed patch aggregation.*

We demonstrate the practical use of local inspection in experimental images by predicting the quality of imaged objects while approaching and reaching the optimal focus. Figure 5 shows the patch quality and patch weight for bovine pulmonary artery endothelial cells (BPAE) mitochondria measured at different focal positions. The three imaged cells lie at slightly different focal positions and are not in the optimal focus simultaneously (Figure 5 (a)). This characteristic is successfully identified by the patch quality, as shown in Figure 5 (b): some structures of the upper-left cell reach a good focus already in the second image (-2), with optimal results reached in the -1 and 0 positions, where also the lower cell is in optimal focus. Instead, few structures of the middle-right cell reach the focus in the -1 position, and gradually leave space to different regions of the cell until the +1 position. A similar behaviour can be noticed for the patch weight. Notably, for high quality images the low-quality score of the background region does not contribute to the overall quality prediction thanks to the low patch weight, as shown in Figure 5 (d).

To validate further the flexibility of the local quality estimation, we include in Supplementary Figure 2,3 the predictions of semisynthetic images with noise and blurring varying linearly within single FOVs.

These results highlight the differences between MM-based methods, which are reliable for the overall detection of artifacts (provided that the underlying markers can be correctly computed), and patch-wise DL approaches, which inspect the artifacts locally and can be used to identify high-quality regions.

# 3. Conclusion

In this paper, we presented µDeepIQA, a DL-based method for automatically predicting image quality in optical microscopy, adapted from an established method for natural images.[10, 13] We retrained and applied an architecture comprising a CNN-based feature extraction path and two parallel regression branches to predict single-patch quality and weight (Figure 2). Supervised training was performed on a semisynthetic dataset incorporating real experimental images, which can be expanded and tuned thanks to the simulation of experimental artifacts and sample structures. The applicability of the model has been demonstrated in two experimental scenarios: predicting the image resolution and a comprehensive quality score. Thanks to the parallel regression branches, the predicted value is a weighted average of the patches, calculated by assigning a patch saliency that depends on the patch content. This approach evaluates high-quality images excluding background regions with no relevant content, but flexibly re-weights all patches when uniform artifacts or dense structures are present.

While this approach has been originally designed for natural images, we have demonstrated significant advantages for optical microscopy studies. µDeepIQA shows replicable performance compared to standard IQA methods based on established quality metrics when applied to semisynthetic datasets with standard experimental artifacts. In addition, when applied to artifact combinations not included in the training dataset, the model can generalize and predict robust quality rankings with good correlation with the real quality. For example, µDeepIQA regularizes instable behaviour of standard methods in noise-free images with high smoothing and vignetting. This might be of particular interest when evaluating denoised images.

Another interesting feature of µDeepIQA is the local quality prediction down to single patches (32 x 32 pixels, in this paper). Patch quality and weights are originally utilized as internal tool to compute image predictions as a weighted average, where single patches are evaluated independently and contribute to the overall quality estimation. However, patch predictions provide local inspection of the images because they represent the actual patch content. In this way, the model identifies high-quality regions of interest in microscopy measurements, which often feature large background regions and variable artifacts within the FOV. On the other hand, if specific artifacts must be detected in full images, MM-based approaches still provide a straightforward and reliable global prediction.

Along the same direction, examining single-sample prediction reveals that µDeepIQA performs better for larger structures that are localized in specific regions of the images. Predictions are more challenging for filamentous, thin structures that are uniformly distributed across the FOV and do not create local intensity changes. This suggests that a minimal structure size is beneficial for DL prediction although, notably, performance difficulties affect mostly simulated structures.

In addition, once the training process is finalized, µDeepIQA provides substantial acceleration of IQA compared to the computation of standard metrics and metrics-based modelling, thus becoming advantageous for experimental applications.

In conclusion, µDeepIQA shows the potential of DL-based methods for fast, robust IQA with patch-wise resolution. The advantages listed above are of great interest for implementing experimental pipelines in biomedical studies, where improving and validating the generated images is essential for reliable data analysis and diagnostic results. In this context, µDeepIQA has demonstrated its potential to be a practical tool for assessing image quality quickly and flexibly for a variety of image artifacts, surpassing the need for a task specific evaluation method.

# 4. Methods

## 4.1 Model architecture

The CNN is adapted from a previous work[13] inspired by the VGG architecture[14] to evaluate single channel images acquired by optical microscopes (Figure 2 (a)). The training is executed for the target parameter to be predicted, either a quality marker such as the image resolution or a comprehensive quality score, thus generating a collection of models tuned to specific target predictions.

The input of the CNN are patches of size 32 x 32 pixels extracted from single-channel images. The images are rescaled from 0 to 1 and the value of single patches is not re-normalized to maintain the relative intensity levels of dark or bright image regions. Each patch passes through a feature extractor made of a series 5 pairs of convolutional layers (Figure 2 (b)) with 3x3 kernel and increasing number of channels and a final max pooling layer of 2x2 kernel, organized in the following cascade structure: conv3-32, conv3-32, maxpool, conv3-64, conv3-64, maxpool, conv3-128, conv3-128, maxpool, conv3-256, conv3-256, maxpool, conv3-512, conv3-512, maxpool. The generated feature vector has size 1x512 and is sent to two parallel branches of fully connected layers (FC). Here, a sequence of one FC-512 and one FC-1 layers regress the features and generate one number. FC layers have dropout with 0.5 ratio. The first branch of FC layers is used to estimate the quality of the patch $y_i$, while the second branch is used to estimate the weight of the patch $\alpha_i$ on the overall image quality prediction $y$. CNN layers have ReLU activation and output size equal to the input size. FC layers have linear activation, except for the FC-1 layer of the weighting branch, which has ReLU activation plus a small function set to 1e-6. This architecture results in 5 237 986 trainable parameters.

The final image quality $y$ is obtained with a weighted global estimation of the $N_p$ patch predictions:

$$y = \frac{\sum_{i=1}^{N_p} \alpha_i y_i}{\sum_{i=1}^{N_p} \alpha_i} \tag{1}$$

## 4.2 Loss function

The loss function $E_{wp} = E_w + E_p$ is designed to optimize the patch-wise estimation and the weighted image estimation with the following expressions:

$$E_w = \left| \frac{\sum_{i=1}^{N_b} \alpha_i y_i}{\sum_{i=1}^{N_b} \alpha_i} - q_t \right| \tag{2}$$

$$E_p = \frac{1}{N_p} \sum_{i=1}^{N_b} |y_i - q_t| \tag{3}$$

Here, $E_w$ minimizes the mean absolute error (MAE) between the weighted image estimate and the true image label $q_t$, while $E_p$ minimizes the MAE between the single patch estimate and the true image label. Here $N_b$ indicates the number of patches in each training batch composed of patches from a single image.

## 4.3 Training

The CNN is trained for 300 epochs on a dataset composed of semisynthetic images of size 512 x 512 pixels. The batches contain 128 patches coming from the same image, meaning that each image contributes to two training batches. The model is trained with Adam optimizer and learning rate of 1e-4 and is validated on the mean squared error $E_w$ between the weighted patch aggregation of the batch and the true image label. Batch size and learning rate are chosen to optimize the network performance and the training time (see Figure 6).

Training, validation and test dataset are composed of semisynthetic images, which are generated from 6 custom simulated structures and 9 experimental samples measured by confocal and two-photon excited fluorescence, which are taken by the Fluorescence Microscopy Denoising (FMD) dataset[15]. Supplementary Figure 1 shows an overview of the experimental and simulated samples. The original images are degraded by 5 simulated experimental artifacts, inspired by a previous work[1]: blur, 3 kinds of mixed Poisson-Gaussian (MPG) noise, and vignetting. In addition, a low level of noise is applied to generate one high-quality version of the images and is also introduced in the blurred and vignetted images to mimic experimental measurements and ensure correct computation of the state-of-the-art quality metrics. Blurring is simulated as a convolution with a Gaussian point spread function of varying standard deviation $\sigma_{blur}$. MPG noise is composed by an additive contribution of three terms: dark noise, a signal-independent background with Poisson distribution $P(\lambda_{dark})$; readout noise, a signal-independent noise with Gaussian distribution $G(0, \sigma^2_{read})$ with zero average and variance $\sigma^2_{read}$; and shot noise, a signal dependent noise with modified Poisson distribution $P(\lambda_{shot}, \alpha_{shot}\, \lambda_{shot}) = P(s, \alpha_{shot} s)$ with average equal to the clean image $s$ and variance modulated by a factor $\alpha_{shot}$. Vignetting is simulated as a multiplication with a maximum normalized Gaussian illumination mask with standard deviation $\sigma_{ill}$. Images are simulated by generating a random value of the free parameter from a normal distribution with the parameters reported in Supplementary Table 1. Possible negative values are removed from the images, then pixel values are rescaled to 0-1.

The training dataset includes 24 fields of view (FOVs) per simulated structure and 16 FOVs per experimental sample, which are degraded by each artifact, resulting in a total of 1728 images. The validation and test dataset includes 3 fields of view (FOVs) per simulated structure and 2 FOVs per experimental sample, which are degraded by each artifact, resulting in a total of 216 images. All final datasets contain the same proportion of simulated and experimental FOVs. Before training, the true labels of training, validation and test images are z-score normalized to the mean and standard deviation of the training dataset for better performance of the training process.

## 4.4 Prediction datasets

### *4.4.1 Noise-free dataset*

To validate the behaviour of the CNN with non-standard artifacts we generated further synthetic and semisynthetic images. We generated one FOV per simulated structure and selected one FOV per experimental sample from the clean images used for the test dataset. We degraded the selected FOVs by adding 5 noise-free increasing levels of blurring and vignetting. 5 independent high-quality images with low noise level are also generated from the same FOVs, to demonstrate that the predictions are stable in absence of artifacts. The values of the corresponding free parameters are reported in Supplementary Table 2.

Kendall's rank correlation coefficient (KRCC) is computed for this dataset by correlating the predicted quality score with the real artifact level (supplementary Figure 2). To obtain a single correlation score for the dataset, the images are grouped according to the real artifact level. Then, the MM-IQA predicted quality score and µDeepIQA-predicted quality score are averaged on all samples within each group. KRCC is computed between the average score and real artifact for each artifact level.

### *4.4.2 Out of focus experimental images*

The images are part of the Parasites datasets published open source by previous works.[2, 16] The dataset includes confocal microscopy measurements of BPAE cells measured at different focal positions, located at 0, ±0.6, ±1.2, and ±1.8 µm from the optimal focal position along the optical axis. We selected the nucleus, mitochondria and actin channels of one sample imaging three different cells at slightly different optimal focus to visualize how the patch predictions change when different regions of the image reach the optimal focus.

### *4.4.3 Images with increasing noise or blur*

Using the same clean FOVs of the test dataset, we generated semisynthetic images with blur or MPG noise linearly increasing from the top to the bottom of the image. The free parameters of the artifacts are increased for each line of pixels as reported in Supplementary Table 3, starting from a value of 0 at the top line.

## 4.5 Generated models

The study mainly describes the results generated by two models trained to predict the FRC resolution and the image quality score. In these cases, the model provides not only an advantage in terms of stability and local prediction, but also a potential acceleration of IQA.

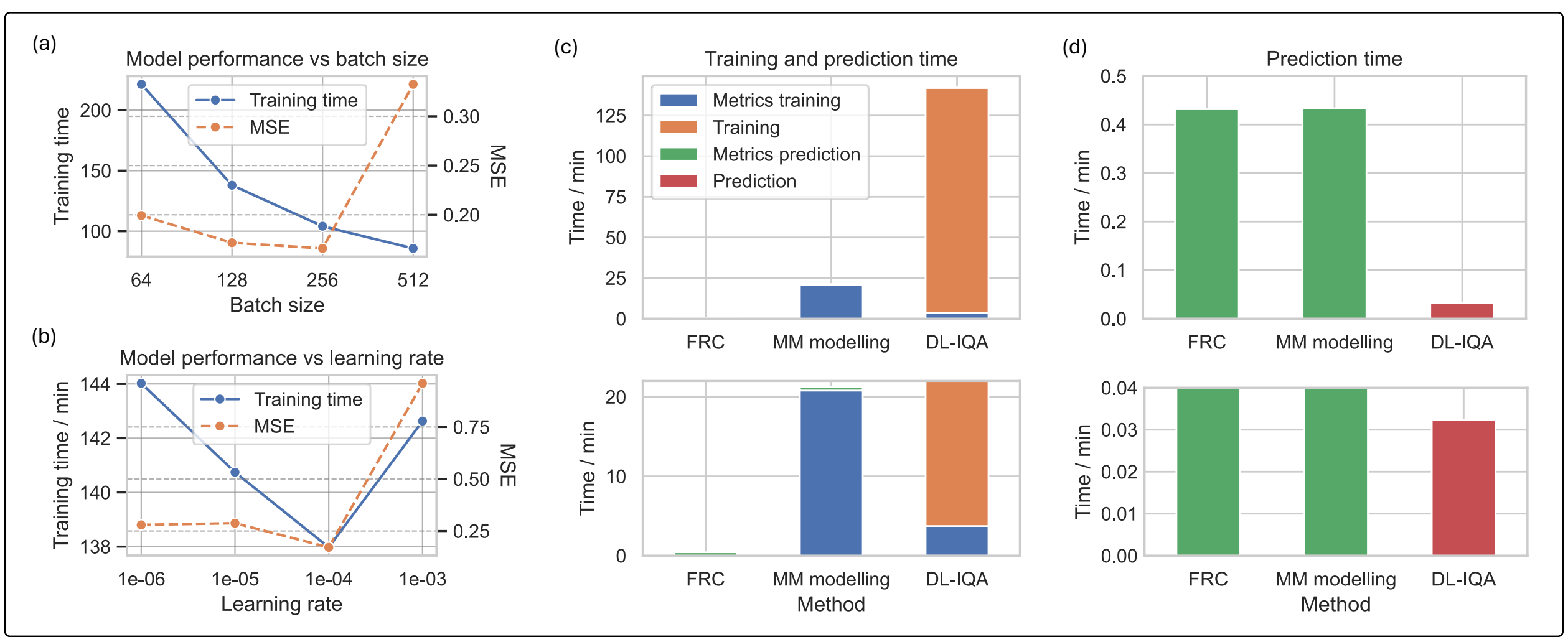

*Figure 6 – **Optimization of learning rate and batch size, and estimation of training and prediction time for metrics-based and DL-based methods.** (a) Training time and mean squared error (MSE) computed on the test dataset as function of the learning rate for the DL model trained on the resolution estimated by Fourier ring correlation (FRC). (b) Training time and mean squared error (MSE) computed on the test dataset as function of the batch size for the DL model trained on FRC resolution. (c) Training and prediction time required for the direct computation of FRC, for the ML-based modelling of the quality metrics, and for the DL method. Four different steps are included: computation of the metrics of the training dataset, model training, computation of the metrics for the prediction dataset, prediction. (d) Prediction time required for the direct computation of FRC, for the ML-based modelling of the quality metrics, and for the DL method. DL methods require longer preparation time due to the training process but provide a considerable acceleration of the prediction time for unknown datasets. ML-based modelling requires negligible training time but requires the computation of the quality metrics for every dataset to be predicted. FRC can be directly computed for the images without prior training, but its computation is significantly slower than the DL prediction.*

### *4.5.1 Prediction of FRC*

The model is trained in a supervised manner, receiving as input the training images and the true FRC resolution computed with the single-image implementation[17].

### *4.5.2 Prediction of image quality*

The model is trained in a supervised manner, receiving as input the training images and, as true label, the quality predicted by Multi-Marker IQA (MM-IQA)[1]. MM-IQA is a machine learning-based approach to predict image quality in fluorescence microscopy from quality markers extracted from the images. MM-IQA generates a classification of image artifacts and its output probability for the reference (i.e., high-quality) class can be used as quality score to rank the images. We used the MM-IQA model trained on the semisynthetic dataset proposed in the original work, without selecting the background region for each image. Then, we predicted the MM-IQA reference probability for the µDeepIQA training dataset and used it as true quality score label. This means that the original quality score is bound to [0,1], with high values assigned to high quality images.

## 4.6 Computational time

We benchmarked the execution time of the CNN to existing approaches on a GPU (Persistence-M, NVIDIA A100-PCIE-40GB) using 3 TensorUnits and 20GB of RAM. The comparison in showed in Figure 6. Here, we estimate the preparation and prediction time to assess 100 unknown images of 512x512 pixels.

The CNN prediction of the FRC resolution requires the computation of the FRC resolution for the training dataset (3.72 minutes), the training time for the CNN (2 h 20 minutes), and the prediction time for 100 images (1.94 s). The standard computation of single-image FRC is rather slow (25.9 s) because it involves the computation of the Fourier transform and an iterative process to compute the correlation at all spatial frequencies.

The CNN prediction of the quality score requires the computation of 7 quality metrics for the training dataset (3.72 minutes), the prediction of the quality score by MM-IQA (10 ms), the CNN training time (2h 20 minutes) and the final prediction for 100 images (1.94 s). On the other hand, the MM-IQA model requires the computation of the metrics at each prediction round (26 s), plus the prediction (1 ms). Moreover, if the MM-IQA must be retrained, it requires 20 minutes to generate the quality metrics of the training dataset. As we can see in Figure 6, the CNN requires a higher but feasible training time compared to state-of-the-art methods, but provides fast predictions and almost no preparation effort when new images must be assessed.

## Acknowledgements

This work is supported by the BMBF, funding program Photonics Research Germany [13N15706 (LPI-BT2-FSU), 13N15719 (LPI-BT5)] and is integrated into the Leibniz Center for Photonics in Infection Research (LPI). The LPI initiated by Leibniz-IPHT, Leibniz-HKI, Friedrich Schiller University Jena and Jena University Hospital is part of the BMBF national roadmap for research infrastructures. Co-funded by the European Union [ERC, STAIN-IT, 101088997]. Views and opinions expressed are however those of the author(s) only and do not necessarily reflect those of the European Union or the European Research Council. Neither the European Union nor the granting authority can be held responsible for them.

## Data availability

µDeepIQA is publicly available at https://git.photonicdata.science/elena.corbetta/udeepiqa. This repository includes the pretrained models used in this publication and all source data to reproduce the results. Training, validation, test and prediction datasets, as well as source data, are available in Zenodo: 10.5281/zenodo.17184938 (the upload will be made public after acceptance of the manuscript).

# Global-to-local image quality assessment in optical microscopy via fast and robust deep learning predictions – Supplementary material

Elena Corbetta[a,b], Thomas Bocklitz[a,b]*

c. Leibniz Institute of Photonic Technology, Member of Leibniz Health Technologies, Member of the Leibniz Centre for Photonics in Infection Research (LPI), Albert-Einstein-Strasse 9, 07745 Jena, Germany.
d. Institute of Physical Chemistry (IPC) and Abbe Center of Photonics (ACP), Friedrich Schiller University Jena, Member of the Leibniz Centre for Photonics in Infection Research (LPI), Helmholtzweg 4, 07743 Jena, Germany

*Corresponding author

E-mail address: thomas.bocklitz@uni-jena.de

# 1. Overview of samples included in training, validation, and test dataset

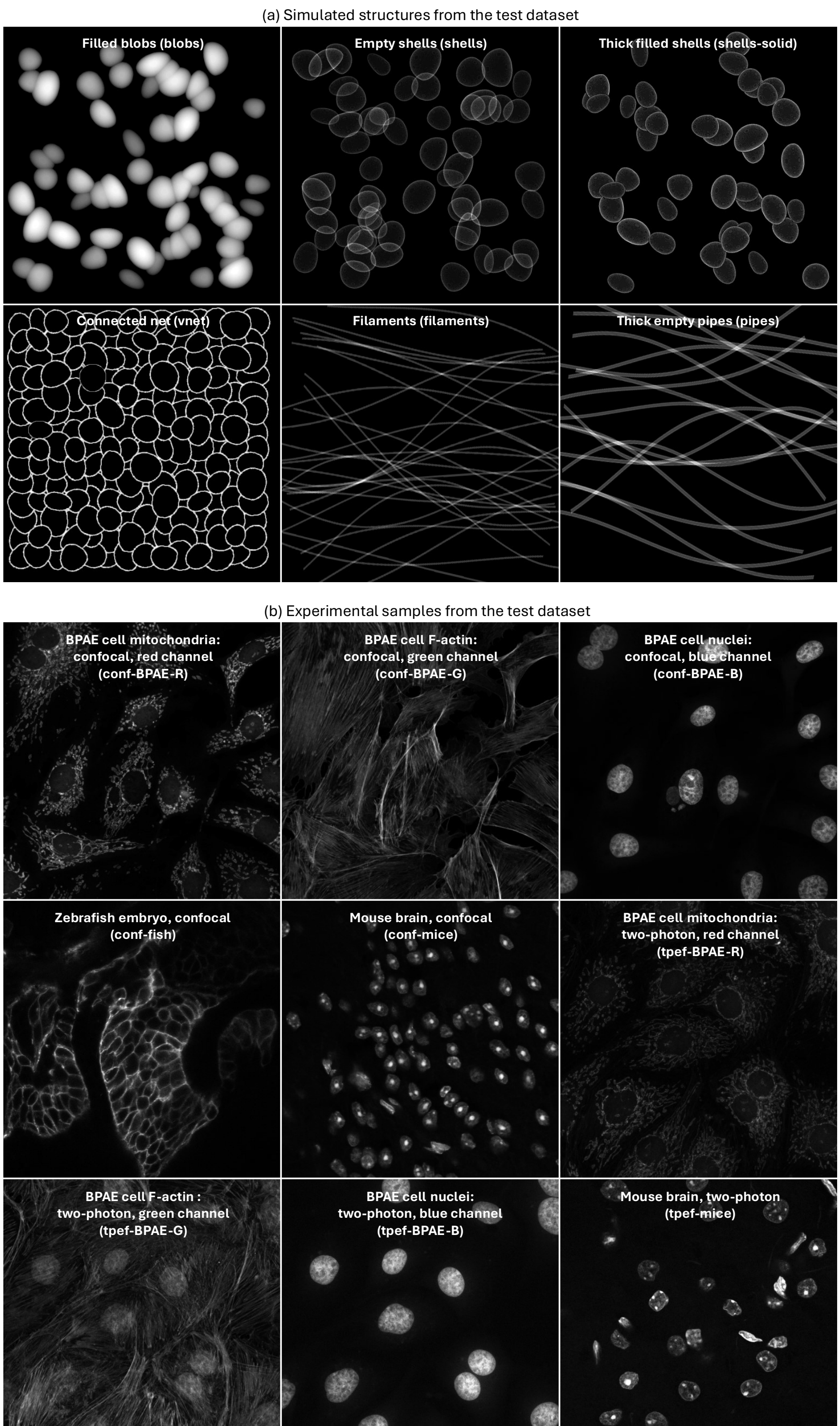

*Supplementary Figure 1 –* ***Overview of clean simulated and experimental samples used to generate training, validation and test dataset.*** *(a) Custom simulated images. Different fields of view are generated by setting parameters for object number, size and orientation and generating random objects. (b) Experimental images taken from the Fluorescence Denoising Microscopy dataset. Size of the experimental images: 153,6 x 153,6 µm. Labels within the brackets indicate the names assigned to the images in the generated datasets according to the imaged sample.*

## 2. Kendall's correlation coefficients

(a) Kendall's correlation: prediction of quality score

| Rankings | µDeepIQA vs Real artifact | MM-IQA vs Real artifact | MM-IQA vs µDeepIQA |
|---|---|---|---|
| **Blurring** | 1.0 | -0.2 | -0.2 |
| **Vignetting** | 1.0 | 1.0 | 1.0 |

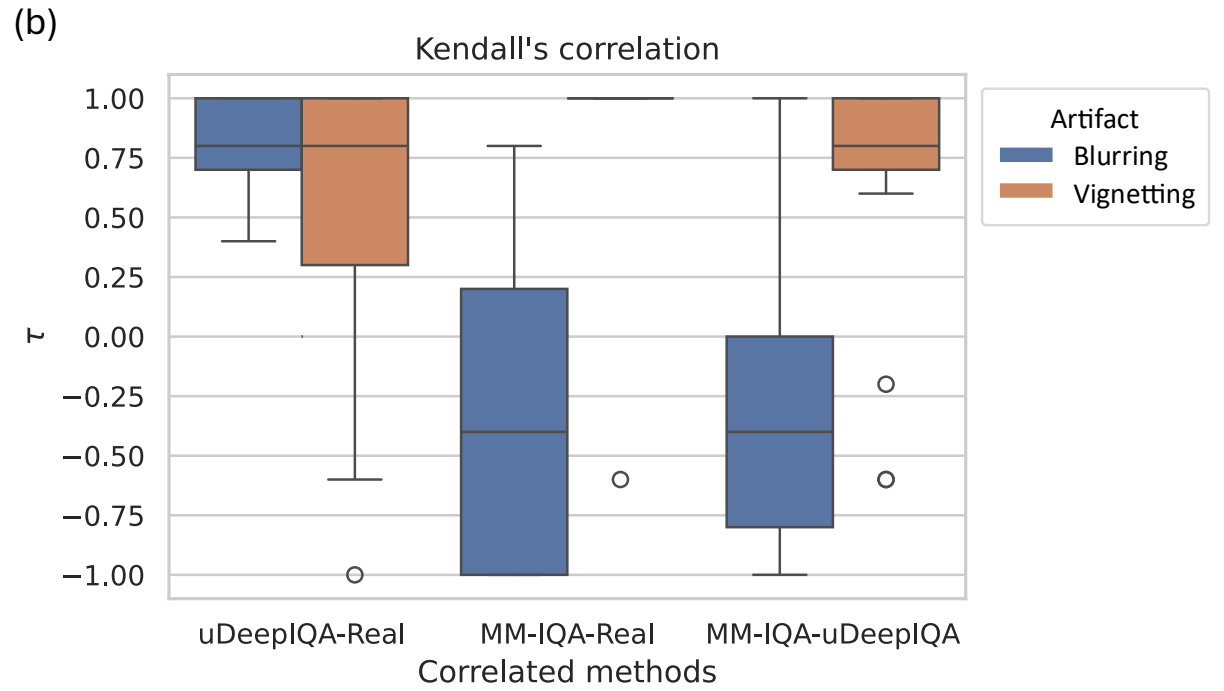

*Supplementary Figure 2 –* ***Kendall's correlation computed for the predictions of noise-free images with increasing blurring and increasing vignetting.*** *(a) Images of the prediction dataset are grouped according to the real artifact level. The MM-IQA predicted quality score and µDeepIQA-predicted quality score are averaged on all samples within each group. Then, Kendall's correlation is computed between the average score and the real artifact level. The table shows Kendall's tau computed between the ranking indicated in the top headers. Optimal correlation can be achieved by both methods for vignetted images, but the quality ranking for blurred images is well predicted only by µDeepIQA. (b) Boxplot of Kendall's tau computed separately for every sample of the dataset with increasing artifact.*

## 3. µDeepIQA patch predictions for spatially varying artifacts

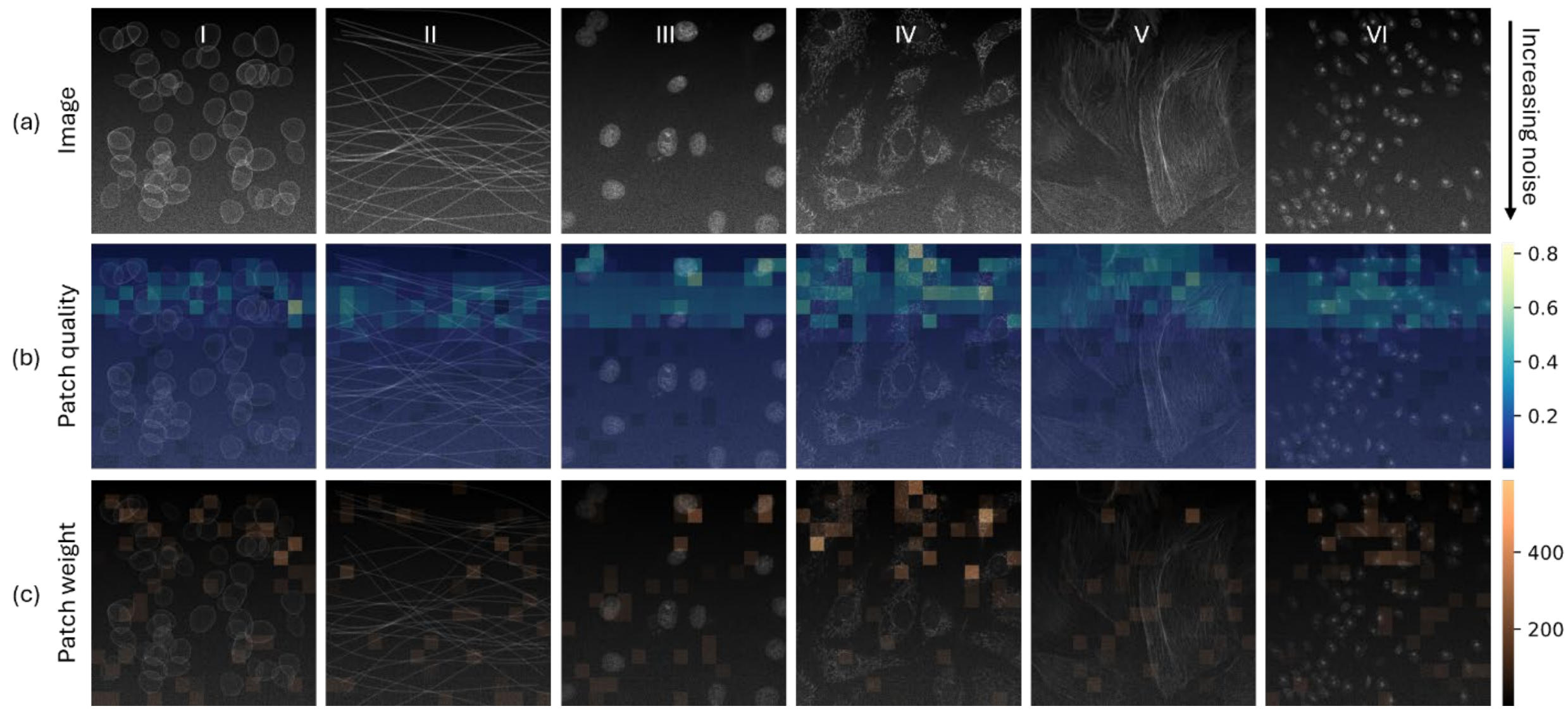

*Supplementary Figure 3 –* ***Inspection of the patch quality and weight predicted by the DL model for semisynthetic images with variable noise.*** *Synthetic (I-II) and experimental (III-VI) are degraded by variable mixed Poisson-Gaussian noise, with increasing image quality along the direction of the yellow arrow (a), as reported in Supplementary Table 3. The predicted patch-wise quality (b) and weight (c) follows the trend of the artifact, with higher and more variable values in the high-quality regions of the images and uniform lower values in the noisy regions. In the high-quality regions, the patch quality varies also according to the image content, with higher values overlapped with the imaged objects, as in III-(b), IV-(b), and V-(b). Patch size of experimental images: 9,6 µm.*

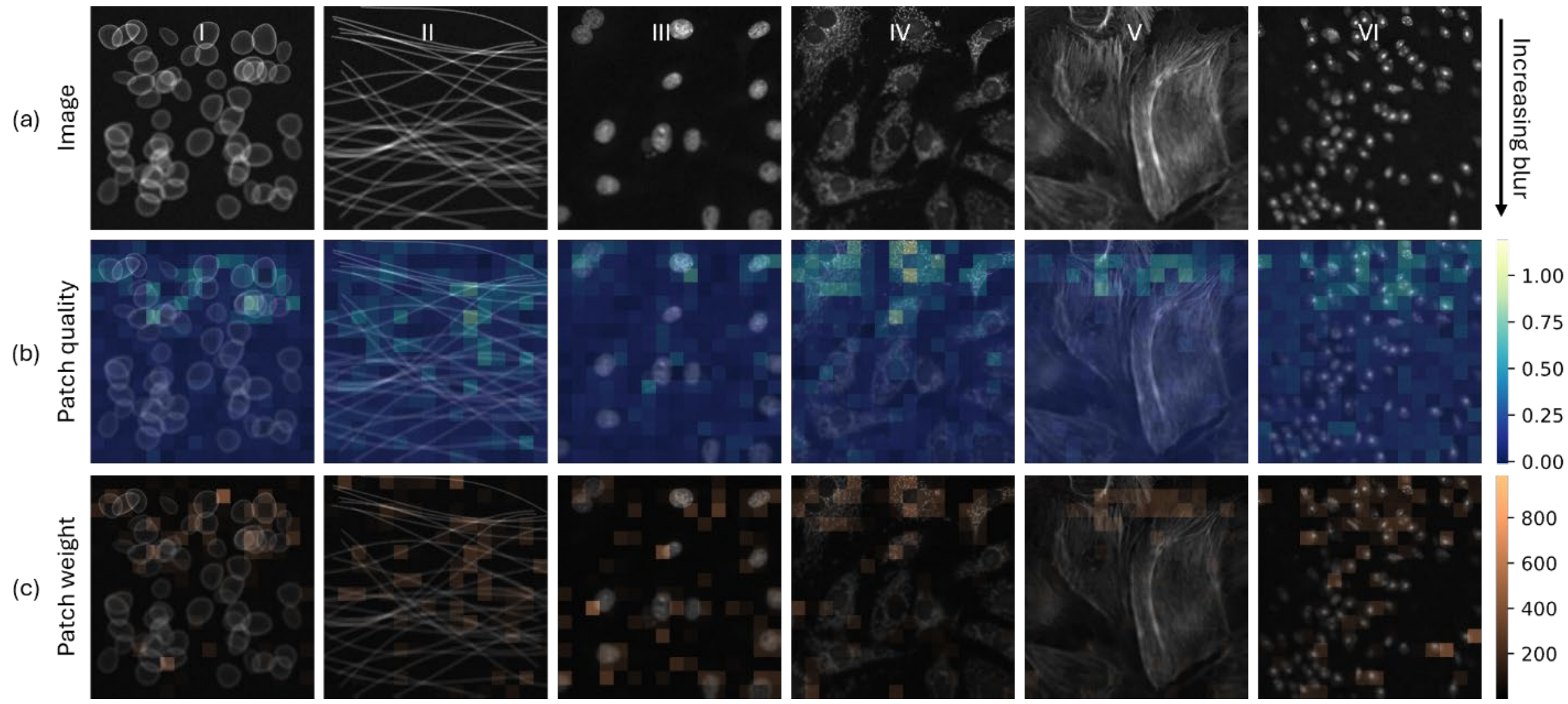

*Supplementary Figure 4 –* ***Inspection of the patch quality and weight predicted by the DL model for semisynthetic images with variable blur.*** *Synthetic (I-II) and experimental (III-VI) are degraded by variable blur, with increasing image quality along the direction of the yellow arrow (a), as reported in Supplementary Table 3. The predicted patch-wise quality (b) and weight (c) follows the trend of the artifact, with higher and more variable values in the high-quality regions of the images and uniform lower values in the noisy regions. In the high-quality regions, the patch quality varies according to the image content, with higher values overlapped with the imaged objects, as particularly visible in III-(b) and IV-(b). Patch size of experimental images: 9,6 µm.*

## 4. Free parameters to generate simulated artifacts

| Artifact | Free parameter | Min | Max |
|---|---|---|---|
| Reference (High quality) | $\lambda_{dark} = \sigma^2{}_{read} = \alpha_{shot}$ | 0,005 | 0,005 |
| Blur | $\sigma_{blur} / px$ | 1 | 8 |
| Dark noise | $\lambda_{dark}$ | 0,3 | 0,5 |
| Readout noise | $\sigma^2{}_{read}$ | 0,7 | 1,5 |
| Shot noise | $\alpha_{shot}$ | 1 | 5 |
| Vignetting | $\sigma_{ill} / px$ | 0,2 | 0,6 |

*Supplementary Table 1 –* ***Free parameters to generate training, validation and test dataset.*** *The table shows the maximum and minimum values of the normal distributions to generate the free parameters in the simulation models. Regardless of the artifact, all images have a baseline of low MPG noise as indicated for reference images. Then, the corresponding free parameter is increased for specific noise types. The parameter for vignetting $\sigma_{ill}$ is multiplied by the lateral size of the image before generating the illumination mask.*

| Artifact | Free parameter | Level 1 | Level 2 | Level 3 | Level 4 | Level 5 |
|---|---|---|---|---|---|---|
| Reference (High quality) | $\lambda_{dark}$, $\sigma^2{}_{read}$, $\alpha_{shot}$ | 0.05 | 0.05 | 0.05 | 0.05 | 0.05 |
| Blur | $\sigma_{blur} / px$ | 1 | 2.75 | 4.5 | 6.25 | 8 |
| Vignetting | $\sigma_{ill} / px$ | 0.6 | 0.5 | 0.4 | 0.3 | 0.2 |

*Supplementary Table 2 –* ***Free parameters to generate noise-free degraded images.*** *The table shows the free parameters for the five increasing artifact levels. Blurred and vignetted images have no underlying noise. The parameter for vignetting $\sigma_{ill}$ is multiplied by the lateral size of the image before generating the illumination mask.*

| Artifact | Free parameter | Min | Increase |
|---|---|---|---|
| Blur | $\sigma_{blur}$ / $px$ | 0 | 0.02 |
| MPG noise | $\lambda_{dark}$ | 0 | 0.001 |
| | $\sigma^2{}_{read}$ | 0 | 0.007 |
| | $\alpha_{shot}$ | 0 | 0.007 |

*Supplementary Table 3 –* ***Free parameters to generate spatially varying artifacts.*** *The table shows the free parameters to generate images with increasing artifacts from top to bottom. The minimum value refers to the top line of pixels in the image. The increase is added for every line of pixels when moving from top to bottom.*